\documentclass[journal=jla,manuscript=perspective]{achemso}

\usepackage[version=3]{mhchem} 

\usepackage{tikz} 
\usepackage{epstopdf}
\newcommand{\orcidicon}{\includegraphics[width=0.32cm]{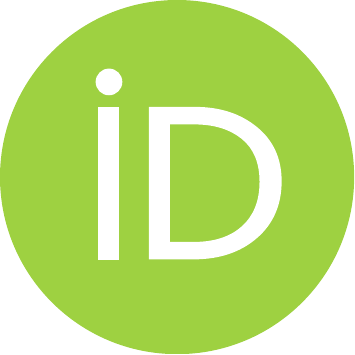}}
\foreach \x in {A, ..., Z}{%
\expandafter\xdef\csname orcid\x\endcsname{\noexpand\href{https://orcid.org/\csname orcidauthor\x\endcsname}{\noexpand\orcidicon}}
}

\usepackage{hyperref}

\usepackage{graphicx}
\usepackage[caption=false, font=normalsize, labelfont=sf, textfont=sf]{subfig}

\usepackage{tabularx}
\usepackage{nameref}
\usepackage{makecell}

\usepackage{soul,color}
\soulregister\cite7
\soulregister\ref7
\soulregister\nameref7
\soulregister\pageref7
\usepackage{xcolor}

\usepackage{gensymb}

\author{Ádám Wolf \orcidA{}}
\email{adam.wolf@takeda.com}
\affiliation{Baxter AG, a Takeda company}
\alsoaffiliation{Doctoral School of Applied Informatics and Applied Mathematics, Óbuda University}
\author{David Wolton}
\affiliation{Takeda Pharmaceuticals International AG, a Takeda company}
\author{Josef Trapl}
\affiliation{Takeda Pharmaceuticals International AG, a Takeda company}
\author{Julien Janda}
\affiliation{Takeda Pharmaceuticals International AG, a Takeda company}
\author{Stefan Romeder-Finger}
\affiliation{Baxalta Innovations GmbH, a Takeda company}
\author{Thomas Gatternig \orcidD{}}
\affiliation{Baxalta Innovations GmbH, a Takeda company}
\author{Jean-Baptiste Farcet \orcidE{}}
\affiliation{Baxalta Innovations GmbH, a Takeda company}
\author{Péter Galambos \orcidB{}}
\affiliation{Antal Bejczy Center for Intelligent Robotics, Óbuda University}
\author{Károly Széll \orcidC{}}
\affiliation{Antal Bejczy Center for Intelligent Robotics, Óbuda University}
\altaffiliation{Alba Regia Technical Faculty, {\'O}buda University, H-8000 Sz{\'e}kesfeh{\'e}rv{\'a}r, Hungary
}

\title
  {Towards Robotic Laboratory Automation \\ Plug \& Play: The ``LAPP'' Framework}

\keywords{Laboratory Automation; Mobile robotics; Autonomous manipulation; System integration; Plug and Play}

\begin{document}

\pagebreak
\begin{abstract}
Increasing the level of automation in pharmaceutical laboratories and production facilities plays a crucial role in delivering medicine to patients. However, the particular requirements of this field make it challenging to adapt cutting-edge technologies present in other industries. This article provides an overview of relevant approaches and how they can be utilized in the pharmaceutical industry, especially in development laboratories. Recent advancements include the application of flexible mobile manipulators capable of handling complex tasks. However, integrating devices from many different vendors into an end-to-end automation system is complicated due to the diversity of interfaces. Therefore, various approaches for standardization are considered in this article, and a concept is proposed for taking them a step further. This concept enables a mobile manipulator with a vision system to ``learn'' the pose of each device and - utilizing a barcode - fetch interface information from a universal cloud database. This information includes control and communication protocol definitions and a representation of robot actions needed to operate the device. In order to define the movements in relation to the device,  devices have to feature - besides the barcode - a fiducial marker as standard. The concept will be elaborated following appropriate research activities in follow-up papers.
\end{abstract}


\section{Introduction}  
The mission of pharmaceutical companies is to provide patients with potentially life-changing treatments as fast and cost-effective as possible. To achieve this, there needs to be continuous focus on reducing operating expenditure throughout the whole organization, from discovery to delivery. However, the time-to-market of new drugs is increasing, along with the cost of development. This is partly due to the fact that it is more complicated to find new target compounds and all of the low-hanging fruit have been picked. In addition, medicines are now becoming specialized and personalized.

This tendency shows a similar - but reverse - curve to the well-known Moore's law. Moore's law explains the exponential evolution in electronics by stating that the number of transistors in an integrated circuit roughly doubles every two years. Eroom's law, on the contrary, states that the number of drugs per billion US\$ is declining logarithmically \cite{Hall2018TheBackwards}. On the other hand, a reproducibility crisis is also observable in R\&D, which means that approximately two thirds of academical research fails to be repeated by the very few peers who decide to take up the non-rewarding task of reviewing and repeating experiments \cite{Mobley2013AClinic, Baker20161500Reproducibility, Ioannidis2016WhyUseful}.

To counter these effects, the pharmaceutical industry is starting to utilize cutting-edge digital, automation and robotic technologies being used in other sectors. As such, the mechanical and machine-manufacturing industries have traditionally been one of the first application fields to utilize  industrial revolutions. Arguably, the fourth (or according to some sources, already the fifth) of these revolutions is currently happening \cite{PlattformSpecifications, Demir2019IndustryCo-working}.
 Cyber-physical systems and the increasingly connected and integrated value chain landscape is transforming industries once more, and the pharmaceutical industry is finding it challenging to keep up. Although the requirements and boundary conditions differ between these sectors, there are many lessons to be taken. Synergies with special industries would suggest that technologies can be adapted. As such, requirements regarding clean-room environments in semiconductor manufacturing, food safety and technological similarities to process industries can be mentioned. Also, the general problem of equipment manufacturers' difficulty to standardize their control interfaces is not unique to the pharmaceutical industry. However, in pharmaceuticals additional requirements, such as good manufacturing practice
 (GMP), regulation is mandatory, working against the desire for flexibility. Strict and complex validation processes make adapting new technologies slower and expensive, and, as a result, the cost/benefit ratio is different to the less regulated industries. One final thing to note is that static robots have only a limited application in pharmaceutical manufacturing, whereas the recently developed mobile industrial robots have the potential to take on a greater percentage of routine and repetitive tasks that tend to be dispersed throughout pharmaceutical facilities and laboratories.

There are multiple factors that are limiting the level-increase
 of automation in pharmaceutical laboratories. Although many laboratory devices represent a high level of automation as self-contained units, the general intention of vendors to provide standardized control interfaces is yet to become universal. Also, mechanical interfaces that would enable an independent manipulator to load samples into the device are not commonly present. Oftentimes, sample-loading takes place through operating complicated door, drawer, snap-in and screw-in mechanisms. This means that most of the devices are still designed for manual operation, i.e. to load the samples, set the parameters, start the process and unload after completion by a human operator.
Although standardization of laboratory consumables is present (e.g. with the SBS-microplate), covering more materials (including tubes, containers and packaging) would further simplify the landscape. This situation needs to be highlighted by the pharmaceutical community so that the vendors can react and make purposeful modifications to enable robotization.

This perspective
 paper provides an overview of relevant technologies that originate from other fields and can benefit pharmaceutical companies; the focus is put especially on the use of mobile robotics in R\&D laboratories. The paper then goes on to discuss the current landscape of laboratory automation, including the state-of-the-art equipment. Finally, it provides a concept proposal aimed at taking the technology to the next level. The goal of the concept is to make the system integration of mobile robotics easier by providing no-configuration, no-teaching, Plug \& Play functionality enabling the cost-effective integration of automated laboratories. The Plug \& Play concept utilizes existing technologies/standards and also aims to incorporate emerging initiatives. The input and support of laboratory original equipment manufacturers
 (OEMs), system integrators and  pharmaceutical companies will be needed for this initiative to succeed; however, the potential applications go well beyond pharmaceutical R\&D laboratories.


\section{Industry Agnostic Evolution of Automation Systems}
This section provides an overview of the established and emerging technologies in other fields, such as the mechanical manufacturing, process- or automotive industries. Analogies and synergies will be highlighted along with naming some areas from where ideas could be adapted or have already been adapted to the needs of laboratory automation. As such, the classical vertical integration approach will be discussed, reviewing the different layers of industrial automation systems. This approach has already been adapted to laboratory automation, as described in section \nameref{sec:labaut}. However, the need for an intelligent infrastructure in combination with connectivity to different assets can only be achieved with new horizontal and distributed integration approaches, such as the internet of things
 (IoT) or the Reference Architectural Model Industrie 4.0 (RAMI). 

With an intelligent infrastructure, the aim is to move from an object- to a service-oriented architecture to reach the full potential of internet of things and services. Therefore, it is a prerequisite to define communication structures and to develop a common language. This is provided as a solution by the Reference Architectural Model Industrie 4.0 \cite{Heidel2019TheVERLAG}. The model ensures that all participants involved understand each other in the course of integrating a robot according to the principles of Industrie 4.0 (I4.0). RAMI 4.0 combines all elements and IT components - including data integrity, privacy and security - in a layer by breaking down complex processes into easy-to-grasp packages. The model helps to transfer the old world of a hardware-based structure and hierarchical communication into the new world of flexible systems with communication among all parties involved.

To get an overview of some of the building blocks of modern digitalized ("smart") industrial production systems, the different levels and layers of planning, control, supervision and data collection are considered. In this regard, many similarities and analogies can be drawn between industrial production and laboratories \cite{Wolf2020DeviceAutomation}. On the highest level, some type of enterprise resource planning
 (ERP) has to take place, while underneath a distributed control system
 (DCS) or a supervisory control and data acquisition system
 (SCADA) resides. These systems feature several layers of scheduling, control and supervision ranging all the way down to the individual machines as the elemental components of the system.

End-to-end digitalization and equipment integration is a crucial aspect in industrial automation and in the process industry. To address these, multiple endeavors took place to provide standardized protocols. One of these, the Open Platform Communications - Unified Architecture (OPC UA), was introduced in 2008 and since has become a worldwide standard \cite{UnifiedFoundation}. OPC UA implements a client-server communication with industrial equipment and systems for control and data collection. As an open standard with an increasingly complex specification, it enables vendors and organizations to model their own control and data structures into an OPC UA name-space. As such, upon the infrastructure, information models provide a specific layer for industry standards and vendor information. The OPC Foundation is closely working together with industrial societies, such as the International Society for Pharmacoepidemiology
 (ISPE), the German Mechanical Engineering Industry Association
 (VDMA) and Zentralverband Elektrotechnik- und Elektronikindustrie
 (ZVEI). The latter is hosting a big number of working groups for defining Companion Specifications for various domains \cite{MarketsFoundation} whereas the ISPE Pharma 4.0 has the focus to add industry and regulatory specific elements that are required for GMP manufacturing of pharmaceutical products.

Based on the RAMI model and intelligence infrastructure, the I4.0 Asset Administration Shell (AAS) concept was developed. It provides a structured interface for connecting I4.0 to physical things by encapsulating all data and information about the assets (active and passive). According to AAS, each physical asset, such as robots, sensors, analytical devices and even workpieces, have their own admin shell representation. Assets can be organized in multiple layers of thematic groups, which have their own corresponding admin shell \cite{PlattformSpecifications}. A similar approach can be applied for laboratory automation, where samples are considered to be the most important assets. Besides, reagents, buffers, consumables and also tools have to be tracked both for audit and for optimization purposes. Several different technologies can serve this purpose, such as traditional barcodes, QR codes, RFIDs, or NFC tags. With new technologies on the other hand, besides identification, providing additional information of the asset is achievable. As such, IoT tags can feature environmental or location sensors, such as indoor GPS or Bluetooth Low Energy technology (BLE) triangulation.

\subsection{Collaborative robots}
On the physical side of automation systems, robots and peripherals play a crucial role in material flow between individual stations. Components often need to be operated in a mixed mode by both humans and robots. Therefore, the applied robotic equipment have to meet certain safety requirements, which are, however, not fully defined yet. In contrast to traditional industrial applications, where robots were separated from humans by physical barriers (cages), cobots are capable and allowed to work in a shared human-robot workspace. Collaborative robots have to avoid collisions with humans or at least minimise the collision forces, as specified in the technical specification ISO/TS 15066:2016 \cite{2016Robots15066:2016}. To achieve this, multiple technical solutions exist, ranging from torque sensing and limitation through sensitive robot skins to monitoring the environment with external sensors.
 When it comes to mobile robots, the safety requirements and regulations are even less established as for cobots. ISO 19649:2017 \cite{2017Mobile19649:2017}
 provides a general vocabulary, whereas specific application fields are covered in separate standards (e.g., ISO 13482:2014 for personal care robots \cite{2014Robots13482:2014}). ISO 18646-1 to 3 \cite{2016Robotics18646-1:2016, 2019Robotics18646-2:2019, 2021Robotics18646-3:2021}
 covers performance criteria and related test methods for service robots, ranging from locomotion through navigation to manipulation.

Autonomous guided vehicles
 (AGVs) have been present in logistics and manufacturing for a long time: they transport packages in warehouses and work-pieces in factories. They vary from heavy-weight autonomous forklifts to smaller AGV platforms capable of docking onto carriers or transporting objects that external mechanisms place on them. They perform autonomous navigation with the help of external navigation aids, such as magnetic or optical stripes or reflective markers.
However, complex collaborative and mobile robots are now emerging on to the market replacing the AGV platforms. These new systems - termed autonomous mobile robots (AMRs) - do not rely on purpose-mounted external references. Instead, they use their on-board sensors to perceive their surroundings \cite{Choset2001TopologicalLocalization}.

Simultaneous localization and mapping (SLAM) technologies utilise odometry to estimate the robot's position. To cope with the drifting effect of integrating velocity measurements over time, feature-points are tracked in an egocentric coordinate-system, e.g. by the means of a light detection and ranging (LIDAR) \cite{Cadena2016PastAge, Bresson2017SimultaneousDriving, Szell2019SLAMEnvironment}. To further decrease uncertainty, sensor fusion can be used, where data from other modalities, such as inertial measurement units (IMUs), are combined. A-priory maps can serve as a basis for SLAM-based navigation, but the generation of maps "on-the-go" is also possible. By the means of constant monitoring of the environment, dynamic obstacles that were not present when the map was created (i.e. objects, other robots and humans) are also detected and can be taken into account at path planning. For safety, close-proximity sensors and bumpers can be hard-wired to stop the robot if a collision is anticipated.

With the new accuracy and flexibility of the AMR, it was a logical progression to equip the basic AMR with one or more robotic arms. This opens up many more opportunities for their deployment. For the purpose of this article, the term mobile manipulator (MoMa) will be used in order to distinguish between the two types of robots. This means not only that external mechanisms are not needed to place the object on the platform, but also that they can interact with their surroundings in a more versatile way. At present, these robots are light-weight and can work in close proximity and collaboration with humans. In contrast to arm-less AGVs and AMRs, MoMas do not need such a high level of structured environment, but they can also operate equipment designed for humans, thus providing a greater scope of operation. This is especially important in laboratory automation (See \nameref{sec:labaut}).

\section{Specifics of Laboratory Automation} \label{sec:labaut}
In this section, the special aspects of automated laboratories will be discussed, and state-of-the-art technologies will be presented. An extra focus will be placed on how these technologies originate from other fields and what comes to their adaptation.

If a process is highly repetitive, a high throughput can be achieved by adapting the components to that specific set of tasks \cite{Konstantinidis2018, Altekar2006AssayApproach}. This means that the whole environment can be optimized for automation by disregarding human ergonomic factors and prioritizing automated material and information flow. In a so-called lights-out operation, no humans are present in the space, be it a warehouse, a factory or a laboratory \cite{RumfordWalker1977TowardMachines}. This vision, however, cannot be achieved with the current state of technology in most cases. The most suitable venue for the implementation of this would be in a facility where routine tests are conducted, for example in a healthcare diagnostics laboratory or in quality control
 (QC) \cite{Burckhardt2018, Croxatto2016LaboratoryChoose}. The newly emerging concept of cloud labs also leverages these approaches with the aim of democratizing resources, just as it was achieved by cloud computing.

On the other hand, in laboratories where higher flexibility is needed, a hybrid human-robot operation is desired. Many tasks, for example measuring out a certain amount of powder from a container, require a level of dexterity that is not yet achievable by robots. Even though many laboratory devices are already automated in a stand-alone fashion, most of them are not optimized for being integrated into an overlaying automation system \cite{Fleischer2018b}. In most cases, they are still designed mainly for human operation with operator interfaces such as touch screens and keyboards \cite{Chu2015}. This means that human workforce is still needed, e.g., to load the samples into the system and also to transport them from one device to the next one.

Similar to manufacturing operations, automated laboratories also feature control systems of various levels . The highest level that will be considered in this context is the functionality of order fulfillment, which enables customers to submit their inquiries for certain tests to be conducted on their samples.
The orders are then processed in a fully- or semi-automated fashion and fed into the control system of the laboratory platform. Worklists are generated and deployed to a so-called scheduler system, which assigns the tasks to different devices \cite{Konstantinidis2018}. The orchestration of each device can be managed trough a number of ways ranging from proprietary vendor-specific protocols to  standard communication channels. Robots play a crucial role in physically connecting the standalone modules of the system \cite{Liu}. A control system has to be capable of assigning the appropriate transportation and manipulation tasks to each robot, similarly to how the tasks of the investigative laboratory procedure (assay) are assigned to liquid handlers or analytical devices. A robot's own control system then interprets the task and specifies the necessary movements to fulfill them. These movements may be defined by manual teaching or by autonomous dynamic pose detection and path planning approaches.

\subsection{High-level Process Control and Process Representation}
To understand a MoMa's role in a laboratory, first the aspects of the high-level process control have to be discussed. An exemplary assay may start by receiving and storing the samples, after which various sample preparation steps follow - such as dilution, mixing with reagents, filtration and centrifugation. After this, incubation, separation and purification may take place, followed by analysis, e.g., by high-pressure liquid chromatography and mass-spectrometry
 (LC-MS), electrophoresis
 or photometric measurement. For prescribing the sequence of steps, various ways of process representation exist. Most of these solutions feature some type of graphical representation similar to a flowchart. Device vendors \cite{MomentumSoftware, SAMICoulter, LucullusWebsite}, system integrators \cite{GreenBiosero, PlateButlerServices, OverlordAutomation, SoftlinxAutomation} and software companies \cite{UniteFlowAG, DynamicScheduler} each provide proprietary scheduler software.

A universal way of process representation comes from business process modeling in the form of the Business Process Model and Notation (BPMN) standard. The software company UniteLabs provides a solution that enables the creation of laboratory workflows in the form of BPMN-diagrams and utilizes the Camunda Process Engine to deploy and run these directly on the laboratory equipment \cite{UniteFlowAG}. The Camunda BPMN Workflow Engine is suitable not only for process representation in the form of diagrams, but is also capable of orchestrating humans, (micro-)services, IoT devices or robotic process automation
 (RPA) bots. This is enabled by API connectors and a Java interface, the latter of which is also used by UniteLabs to implement SiLA commands for device control (see \nameref{sec:com_standards}
 \cite{Feature_definitions/ch/unitelabs/robotGitLab}.

\subsection{Instrument Communication and Control Approaches} \label{sec:com_standards}
Apart from BPMN, another layer of business process automation can also be utilized in laboratory automation: RPA approaches can be used to interface with equipment that do not have APIs \cite{Wolf2020DeviceAutomation}. Automated interaction with a device software's user interface
 (UI) can be achieved by RPA scripts. A basic implementation in C\# was demonstrated by Chu et al. when they used simulated mouse-clicks and keyboard inputs to interface with an LC-MS system \cite{Chu2015}. UI manipulation in a Windows-environment is possible with AutoIT \cite{HomeAutoIt}, whereas UiPath provides a cross-platform RPA tool \cite{AutomationUiPath}.

If a device does have an API, many system integrators would develop their own proprietary drivers to integrate the devices. However, this results in a complicated, non-standardized landscape of software and protocols. To simplify interfacing the laboratory devices and the control system, the Standardization in Lab Automation (SiLA) consortium aims to provide a common communication and device control protocol in automated laboratories. The organization maintains a standard that defines laboratory devices as servers with a predefined set of features that a client can call. Each SiLA-feature has a defined set of parameters and properties and can be implemented in various programming languages, such as Python, C++, C\# and Java. In this manner, both device vendors and system integrators have the means of making their products capable of standardized interoperability. As such, there is already a feature definition for mobile manipulators, which - among others - contains services for battery control, calibration, gripper control, robot control and teaching \cite{Feature_definitions/ch/unitelabs/robotGitLab}.

With a similar aim, the Laboratory Agnostic Device Standard (LADS) initiative was launched in 2020 by the Spectaris German Industry Association for Optics, Photonics, Analytical and Medical Technologies \cite{NetworkedSPECTARIS, LADSFoundation}. Built upon the OPC-UA protocol, a laboratory-specific information model is under development for use cases, such as monitoring, control, notification, program- and result management, asset management and maintenance. Keeping in mind the short innovation cycles typical for laboratories, the development of the associated standards has to be kept equally short by a device-type agnostic information modelling approach. Chosing OPC-UA as a basis is justified by the fact that laboratories are increasingly aligning with industrial technologies when it comes to automation. On one hand, this means that approaches from manufacturing and process industries are being adapted to the specific needs of laboratory automation. Furthermore, laboratories are becoming more and more integrated into the industrial manufacturing and process infrastructure. This is true not only for QC labs, which serve as a feedback loop, but also for development laboratories, which are rolling out newly-developed technologies to manufacturing scale. In both scenarios, an aligned and well-integrated infrastructure is highly beneficial. However, production-near environments, such as QC, are more suitable for LADS, in contrast to SiLA, which is rather aimed at R\&D labs.

Even though there are endeavors for standardizing laboratory system integration, the landscape is still complex with a variety of proprietary and semi-standardized solutions. With the intention to collect integration information for the various devices and sub-systems, the Universal Integration Knowledge Base (UIKB) was founded \cite{LabConfluence}.
Its main goal is to act as a database of successful integration ``recipe'' and APIs, thus saving time and costs during experiments. Based on voluntary data sharing, the community sustained platform is expected to benefit OEMs, system integrators and ultimately the users.

\pagebreak
\subsection{MoMa Technologies} \label{sec:mobrob_labaut}
Fleischer et al. \cite{Fleischer2017} differentiate life science automation systems based on the following factors:
\begin{itemize}
    \item Local distribution of the automation devices (centralized or decentralized)
    \item Flexibility of the automation structure (open or closed)
\end{itemize}

According to their definition, the biggest challenge for system integration lies in decentralized/open systems. Contrary to most traditional industrial applications, where the processes are not changing for a longer period (usually years), in life science laboratories a robot has to cover a range of functions. When acting as a central system integrator, the robot serves as a transportation element. For this, either the devices have to be adapted or the robot has to mimic how a human would operate each device. On the other hand, Fleischer et al. distinguish a flexible robot by the fact that besides transportation it also performs other laboratory tasks, such as sample manipulation \cite{Fleischer2018AnalyticalPipettes}.

According to Fleischer et al. \cite{Fleischer2017}, laboratory robotics solutions can be designed in two general configurations:
\begin{itemize}
    \item Fixed-position robots surrounded by all devices, tools, labware and consumables
    \item Mobile manipulators, which can approach different stations while transporting and manipulating the samples. This approach implements an open automation system
\end{itemize}

Benchtop robot arms either in stationary or rail-mounted manner are now available from multiple laboratory automation system integrator companies. These applications usually focus on handling standardized sample carriers, such as SBS-microplates. The scheduler software controls both laboratory devices and transportation robots, especially starting procedures and monitoring their states. However, such systems require comprehensive integration efforts, in that the placement of the devices around the robots or the tracks has to be carefully designed. In most cases, custom frames are needed instead of the conventional laboratory benches. With this approach, sets of devices can be integrated into an automated island, where the robot serves as a transportation backbone. However, this approach introduces spatial constraints, hence highly limiting the flexibility \cite{Wolf2019a}.

MoMas consist of a mobile base, one or more robot arms and usually a number of sensors, including laser scanners and (3D) cameras \cite{Bohren2011TowardsPR2, Wise, Pages}. These units are capable of navigating to various stations in a facility, picking up objects, such as packages, workpieces or samples, transporting them to the next station and eventually loading them in a device or machine for further processing or storage. Being able to cover a larger area than a fix- or rail-mounted robot, mobile manipulators can increase the flexibility of automated laboratory systems. Their range can cover multiple laboratories across different floors of the same building. For this, interfacing with automated doors and elevators has to be implemented \cite{Abduljalil2019}.

Fleischer et al. \cite{Fleischer2017} consider a MoMa as an integrated robot solution, in that it fulfills both sample transportation and manipulation in an open automation system. To enable the robot to interact with laboratory devices, the interfaces have to be defined. This includes both the communication and control interfaces to be configured and the movements of the robot to be programmed. The latter component calls for an appropriate representation of the movements.

The standard solution for mobile manipulators to locate the graspable objects has become the utilization of fiducial markers
 (FMs) (see section \nameref{sec:fiducials}). This means that the MoMa has to be equipped with a calibrated on-board camera, which is then used to detect the pose of optical markers. To enable this, the markers have to be mounted near the handover positions and the robot motions have to be defined in regards of the marker-defined coordinate frame.

Multiple system integrator companies have already begun to provide MoMas besides the usual stationary or rail-mounted SCARAs as part of their sample transportation solutions. Many of these feature the same type of PreciseFlex robot arm, which has became a standard for bench-top laboratory robotics \cite{PreciseHandler}. As such, the Fraunhofer Institute for Manufacturing Engineering and Automation (IPA) has launched KEVIN, which is based on a Care-o-bot 4 mobile platform, a mounted PreciseFlex arm and a vision system, which is capable of localizing fiducial markers \cite{Traube2019}. A very similar solution is provided by the company Biosero \cite{BioseroLab}. They use the same SCARA, but in this case it's mounted on an OMRON LD90 mobile base, alongside with a different vision system for detecting a different kind of fiducial markers. The solution of UniteLabs and Astech Projects only slightly differs from the above two in the sense that it is based on a mobile platform with a bigger footprint and that instead of a SCARA it features a 6-axis arm from Universal Robots \cite{UniteLabsCompany}. Other than that, the fiducial-based pose detection principle is the same. Moving towards industrial robots, KUKA's MoMa named iiwa has also to be mentioned
 \cite{KMRAG}. This specific robot was utilized by Burger et al. for conducting photocatalysis experiments within a ten-dimensional space \cite{Burger2020AChemist}. It can be concluded that the blueprint of equipping a mobile platform with at least one robot arm and a vision system capable of fiducial-based pose detection is already well established and widespread in the laboratory automation landscape.
The use of MoMas, however, goes beyond the walls of the laboratory when it comes to the pharmaceutical industry. For the Marvin Project of the PM Group and the University College Dublin, an iiwa unit was used for environmental monitoring purposes \cite{AutonomousManufacturing, Ramasubramanian2020OperatorMovement}.

Considering the basic functionality, using an AMR alongside with several stationary robots can serve as an alternative to a MoMa. The two different approaches both have their advantages and disadvantages, and ultimately the specific use-case determines which one proves to be the optimal solution. On one hand, when a higher throughput has to be achieved (e.g. in manufacturing or logistics), setups with conveyor belts and/or stationary robots have to be considered. MoMas might not reach the speed of humans, but they can get closer to their level of flexibility than stationary robots specialized for one certain task. Although a single robot arm might be lower-priced as a complex MoMa, the latter can render multiple stationary units superfluous by fulfilling multiple handling and tending tasks at once. Being able to work 24/7 gives these systems an advantage in comparison to human-based solutions. See Table \ref{tab:comparison} for a comparison. Taking over a wide variety of tasks from humans is also beneficial for minimizing on-site presence during the times of a pandemic, such as COVID. Besides fulfilling transportation and manipulation tasks, MoMas can utilize their wide range of sensors and actuators, which opens up additional possibilities. As such, a MoMa can also be used for remotely monitoring an automated system in a 24/7 operation. Without having to travel to the facility, stand-by personnel can remotely control the robot to navigate to the place where the error occurred. With the robots on board camera and additional sensors, the cause of the error can be evaluated, and, in certain cases, telemanipulation can be used to resolve the error. As a simple example, a freezer door left open can be mentioned, which can easily be closed with a push. Similarly, in the case of the stationary sample-handling robots, a misplaced plate can be pushed back to the hand-over nest - that is if no spillage occurred.

\begin{table}[h]
\centering
\caption{Comparison}
\label{tab:comparison}

\begin{tabular}{|l||c|c|c|}
\hline
& \multicolumn{1}{l|}{Stationary robot} & \multicolumn{1}{l|}{MoMa} & \multicolumn{1}{l|}{Human} \\ \hline

\hline
\hline
Throughput   & High             & Low    & Middle \\
\hline
Availability & High             & Middle & Low    \\
\hline
Flexibility  & Low              & High   & High \\ 
\hline
\end{tabular}

\end{table}

\subsection{Robotic Actions on Marker-based Coordinate System} \label{sec:fiducials}
When it comes to fulfilling handling tasks with any type of robot, precision and accuracy are key considerations. As a reference, it has to be mentioned that robots in laboratory automation are mainly used for automating tasks that were previously conducted by humans. As such, the required precision is hard to define. As a reference, the pipetting and bench-top handling robots can be considered. A popular example of the former, the TECAN Freedom Evo, has 0.5 mm specified for the robotic manipulator arm, whereas the PreciseFlex SCARA arm is given a 0.09 mm value. The precision of a mobile robot on its own lies around 50 mm, which has to be improved by the means of additional position detection methods. Overall, a precision in position of around 1 mm and in angle around 1-2 deg is desired.

As described in section \nameref{sec:mobrob_labaut}, fiducial markers play a crucial part in many of the state-of-the-art MoMa applications. Principally, a fiducial marker is a two-dimensional (usually binary) pattern that can be printed and sticked to
 various surfaces to serve as an optical marker for pose estimation. A computer vision algorithm is used to detect the pattern on a calibrated camera's image and calculate its corresponding position and orientation in relation to the camera coordinate system (frame). One of the traditional applications was augmented reality
 (AR), where virtual objects can be added to a camera-captured scenery dynamically. In these applications, fiducial markers serve as a reference between the physical and the virtual environment in the form of coordinate systems. Garrido-Jurado et al. propose a method for the generation and detection of fiducial markers called ArUco \cite{Garrido-Jurado2014AutomaticOcclusion}. Besides AR, they also mention robot localization as one of the application fields for the technology.

In the discussed applications, fiducial markers are used as user coordinate systems in which the robot motions have to be composed. As such, a user can define the grasping pose in relation to the marker by moving the robot arm to the desired configuration either by hand or by jogging. In other applications, however, robot motions are considered in a modular representation structure. Chu et al. propose the so-called Motion Elements framework specially for the automation of a sample preparation workflow \cite{Chu2016AutomationApplications}. Complex hierarchical motion representation structures are also present in surgical robotics \cite{Nagy2019AAutomation, Vedula2016AnalysisTask}. As a crucial part of the framework proposed in the present article, the concept of Action Primitives will be introduced.

\section{Laboratory Automation Plug \& Play - Concept Proposal} \label{sec:proposal}
As seen, the state-of-the-art solutions for a mobile-robot-based end-to-end laboratory automation have limitations. The following bullet-points list these in the order of severity.

\pagebreak
\begin{itemize}
    \item Lacking standards
    \begin{itemize}
        \item Device and data interface standards (Existing ones, such as SiLA, are yet to become universal
        \item Materials and consumables
        \item Mechanical interfaces
        \item Regulations 
    \end{itemize}
    \item Missing conformity with strict guidelines present in the pharmaceutical industry (e.g. GxP \cite{GAMPEngineering})
    \item High integration effort
    \begin{itemize}
        \item From the side of the system integrator
        \item From the side of the user to adapt their workflows
    \end{itemize}    
    \item IT security concerns
    \item Robots have a limited adaptability
    \begin{itemize}
        \item Dexterous gripping is still considered to be an immature technology 
        \item Identifying and localizing devices is not well-established. Speed and precision could be improved
        \item Flexible robotization in a laboratory automation environment still cannot cope with humans in regards of speed and dexterity 
    \end{itemize}
    \item Devices are not robot-ready
    \begin{itemize}
       \item Self-contained proprietary solutions
       \item Lacking automatic lids and doors 
       \item Complicated insertions of custom carriers   
    \end{itemize}
\end{itemize}

\pagebreak
To overcome the limitations of the state-of-the-art solution for integrating devices in a robotized laboratory, the Laboratory Automation Plug \& Play (LAPP) Framework is proposed. According to this approach, a LAPP-enabled device will have to feature two types of optical tags on the front face of the device:
\begin{itemize}
    \item One 1D or 2D barcode for referencing device information, e.g., in the DMRE Datamatrix format
    \item One fiducial marker for pose detection
\end{itemize}
The former code will enable a robot equipped with a vision system to access an online database (e.g., the UIKB, see section \nameref{sec:com_standards}) and fetch device-specific information, such as represented in Table \ref{tab:fetchable_info}. To implement the communication with the database, modern web-technologies have to be used that are already present in industrial automation \cite{Galambos2020CloudApplications}. As such, websocket, REST-API, Graph-QL and gRPC can be named, the latter of which is also used by SiLA.

\begin{table}[h]
\centering
\caption{Information to be fetched from local systems and the cloud}
\label{tab:fetchable_info}
\begin{tabularx}{\textwidth}{X|p{0.5\textwidth}}
  \textbf{Piece of information to be fetched} & \textbf{Utilization} \\
\hline
\begin{itemize}
    \item Device identification
    \item Vendor
    \item Model
    \item Serial Number 
\end{itemize} &
\begin{itemize}
    \item Asset management
    \item Record in associated systems, e.g., laboratory information management system
 (LIMS)
    \item Track \& trace
\end{itemize}\\
\hline
\begin{itemize}
    \item Maintenance information
\end{itemize} &
\begin{itemize}
    \item Connect to inventory database
    \item Health information
    \item Alarm \& events
\end{itemize}\\
\hline
\begin{itemize}
    \item Live device info
\end{itemize} &
\begin{itemize}
    \item Visualize with mobile devices, AR/MR glasses
    \item Digital twin of the device
\end{itemize}\\
\hline
\begin{itemize}
    \item Device features (e.g., SiLA)
\end{itemize} &
\begin{itemize}
    \item Application programming interface
 (API) definition
\end{itemize}\\

\hline
\begin{itemize}
    \item Action primitives in the coordinate system of the fiducial marker (Global information provided by the device vendor)
\end{itemize} &
\begin{itemize}
    \item Mechanical interaction between the mobile robot and the device
\end{itemize}\\
\end{tabularx}
\end{table}

\clearpage

\begin{figure}[h]
    \centering
    \includegraphics[width=0.8\linewidth]{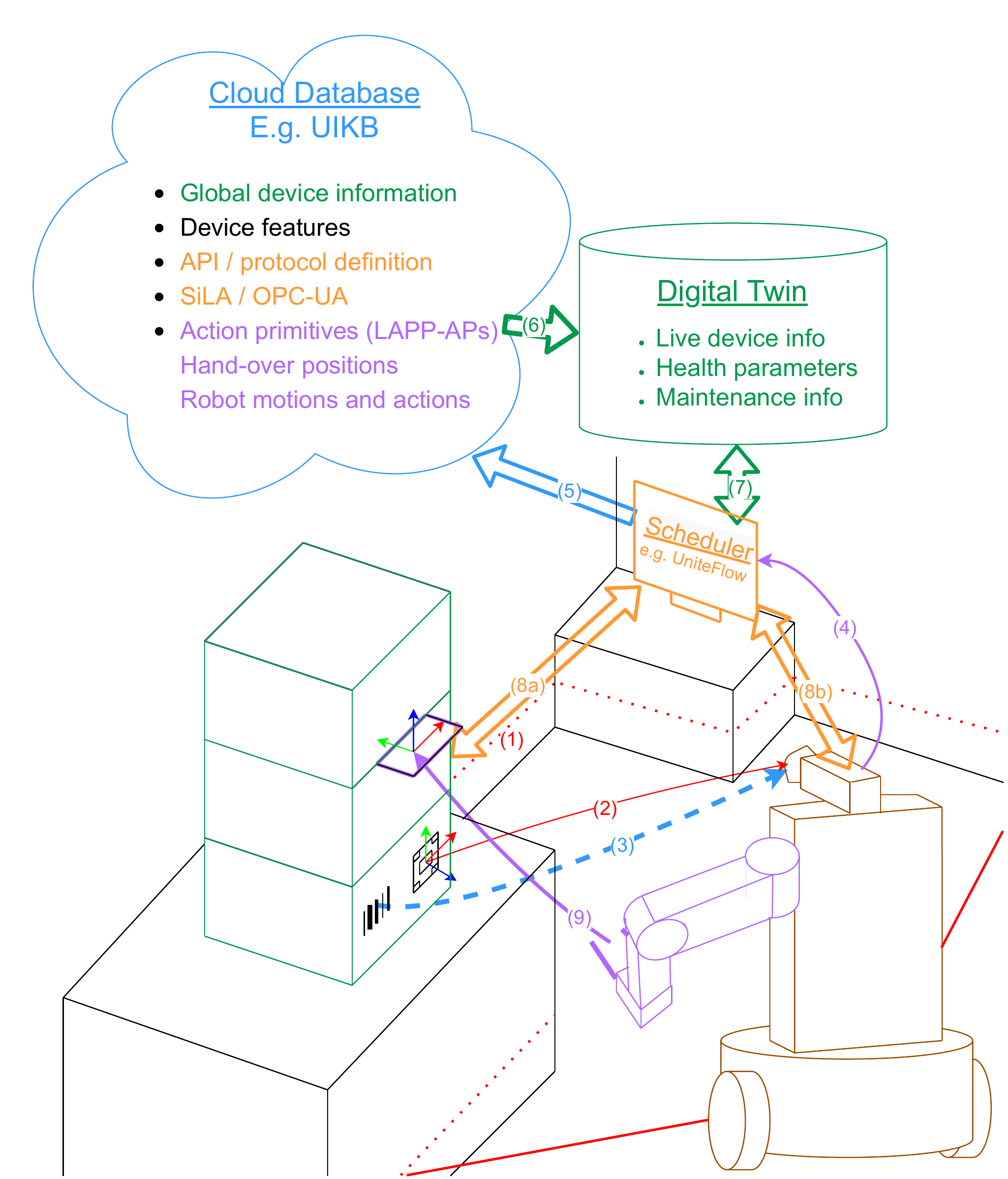}
    \caption{The LAPP sequence (1) Generate map, (2) Detect barcode, (3) Detect fiducial pose, (4) Upload barcode ID to the scheduler, (5) Request device information from the cloud database, (6) Instanciate digital twin from the cloud database, (7) Keep digital twin updated, (8a) Device control, (8b) Robot control, (9) Robotic manipulation}
    \label{fig:concept_overview}
\end{figure}

Utilizing the proposed framework, a mobile robot will be able to operate in a newly installed laboratory environment in a plug \& play manner. This will require an initial, fully autonomous discovery procedure, as Figure \ref{fig:concept_overview} shows:

\begin{itemize}
    \item[(1)]Robot generates the map with SLAM
    \item[(2)]Robot detects device pose with the help of the fiducial marker
    \item[(3)]Robot reads barcode with the vision system
    \item[(4)]Robot uploads barcode ID to the scheduler
    \item[(5)]Scheduler requests device information from the cloud database
    \item[(6)]Scheduler instantiates the digital twin from the cloud database
    \item[(7)]Scheduler keeps the digital twin updated
    \item[(8a)] Scheduler controls the device
    \item[(8b)] Scheduler controls the robot
    \item[(9)]Robot navigates in the laboratory autonomously with the help of the generated map and operates the devices with the help of the fetched action primitives
\end{itemize}
\clearpage

The proposed framework is considered protocol agnostic, meaning that multiple different technical solutions would be suitable for fulfilling the desired functionality. As such, many of the existing building blocks can be utilized, as discussed in section \nameref{sec:labaut}:
\begin{itemize}
    \item High-level process representation and integration in the corporate control architecture can be achieved by a scheduler software, such as \textbf{UniteFlow}.
    \item Interfacing with the equipment can be achieved by means of a communications standard or protocol, such as \textbf{SiLA}
    \item As the central system integrator, a MoMa has to feature a complex robotic middleware, such as \textbf{ROS}, which enables the integration of its on-board actuators and sensors, including the vision system
    \item The vision system has to be capable of detecting the pose of fiducial markers, such as \textbf{ArUco markers}
    \item Robot motions have to be defined in relation to the marker coordinate system. We introduce the term \textbf{Laboratory Automation Plug \& Play - Action Primitives (LAPP-APs)}.
\end{itemize}

As a crucial part of the proposed concept, a LAPP-AP database has to be defined for each laboratory device. This will serve as an extension to the software interface- and feature definitions (e.g., SiLA features) by providing a structural representation for the mechanical interactions needed to operate the device with an external robot (mobile or stationary). Defined in the marker's coordinate system, the motions will be suitable to integrate the device out-of-the-box, in a plug \& play fashion, although, the possibility of manual adjustments and calibration should be kept open. A detailed proposal for the LAPP-AP framework will be elaborated in a separate article (see section \nameref{sec:future_work}).

\section{Discussion, Outlook and Future Work}
\subsection{Impact on the Industry}
To reach the full potential of the proposed framework, an industry-wide involvement is needed. Discussion with OEMs, system integrators and users is crucial for defining the details of the framework in order to provide a standard that is adaptable to a broad range of applications. The standard is as good as its level of acceptance and the extent to which it is actually used. OEMs are already under pressure to make their products capable of a simplified and unified integration by utilizing standards. The present proposal aims to take this a step further - building upon and incorporating existing technologies and initiatives.

Pharmaceutical R\&D process development represents the link between discovery and manufacturing, in that technologies are developed and then transferred. This traditionally constitutes
 the concrete manufacturing processes, which can be of biological or chemical nature; however, engineering aspects, such as system design, are also part of the scope. On one hand, this means scaling up the processes from micro-scale through laboratory-scale and pilot-scale all the way to commercial production. On the other hand, technical solutions that are used in smaller scale laboratories can often be utilized directly or indirectly in higher-scale applications as well. Primarily, various types of laboratories are present not only in R\&D but also in analytics and QC. Throughput and scale may vary, but the ground principles remain analogous. Secondarily, manufacturing itself is becoming more and more personalized, which means that the flexible approaches of a development laboratory may just as well be utilizable. In pharmaceutical manufacturing, a new approach called ballroom concept is gaining importance \cite{Tung2019TheTechnology}. In contrast to traditional manufacturing lines, this enables the dynamic allocation and reconfiguration of automated units. Traditional layouts, such as a conveyor belt being the backbone of the system, cannot fulfill this need. Instead, MoMa-based automated systems are expected to gain importance.

In such a scenario, a robot needs to interact with various types of instruments, devices and machinery, just as in an automated laboratory. Utilizing the proposed plug \& play concept has the potential to simplify the configuration of any system that incorporates a multitude of components and MoMas.

\subsection{Future Work} \label{sec:future_work}
The present article is considered as a high-level overview on the existing approaches, and it also provides an outline on the proposed framework concept. As described in section \nameref{sec:proposal}, a technology agnostic approach was taken, which means that several different implementation possibilities are open. To elaborate these, a series of articles is planned, the first of which will focus on the LAPP-AP concept. In this context, a framework will be proposed to represent laboratory automation specific robot movements in relation to the marker coordinate frame. To enable the robot to operate a variety of devices, the appropriate end-effector has to be identified. Possibilities, such as tool-changers, multi-tools and universal grippers, will be evaluated \cite{Monkman2007RobotGrippers}. 
Following that, the specific technologies for the implementation, including the data structures, database and communication technologies, will be defined with the involvement of the vendors, integrators and users.

\subsection{Summary}
 Applying automation and robotic technologies in the pharmaceutical industry is challenging due to the complexity and highly-regulated manner of its laboratory and production environments. In this paper, various approaches from other industries were discussed, followed by an overview of the state-of-the-art technologies specifically focusing on life science laboratories. Novel technologies, such as flexible MoMas, enable a new wave of automation in these laboratories by connecting the previously standalone components of the system, which were traditionally designed and optimized for manual operation. As such, the integration of these devices has two aspects, one of which is that an overlaying control system has to be able to communicate with them and schedule their operation through control interfaces. On the other hand, to make the process end-to-end automated, physical interaction with the devices is needed, including the transportation of samples, labware and consumables from one station to another. Both of these aspects, however, represent challenges due to the fact that devices from various vendors feature different software and mechanical interfaces. Initiatives for standardization are already present in the industry, but their overall acceptance and global prevalence is not sufficient to effectively enable a Plug \& Play experience when integrating an automated laboratory system. To take the existing standards and integration technologies a step further, the Laboratory Automation Plug \& Play (LAPP) framework was proposed. Being a technology-agnostic approach, it keeps the possibility for multiple possible technical implementations. Ultimately, the concept will enable a MoMa to autonomously ``learn'' the control interfaces and poses (position and orientation) of newly-installed devices. For this, each device will have to feature two types of optical tags: a barcode to reference the integration information from an online database; and a fiducial marker to enable the detection of the pose of the device. One crucial part of the concept is the proposed Action Primitives database (LAPP-AP), which will enable the robot to operate the device in the prescribed manner, including loading it with samples, reagents and labware. For this, robotic actions have to be represented in a structured, modular and parametric fashion. The present paper marks the first of a series, in the course of which the technical details will be elaborated. To reach the full potential of the proposed concept, the involvement of various stakeholders is key. The standardization of such an approach is viable only through an industry-wide discussion including device vendors, system integrators and users.
\pagebreak

\begin{acknowledgement}
This work was funded by Baxalta Innovations GmbH, a Takeda company.

This work was supported by the Doctoral School of Applied Informatics and Applied Mathematics, Óbuda University.

Péter Galambos and Károly Széll thankfully acknowledge the financial support of this work by the project no. 2019-1.3.1-KK-2019-00007 implemented with the support provided from the National Research, Development and Innovation Fund of Hungary, financed under the 2019-1.3.1-KK funding scheme. Péter Galambos is a Bolyai Fellow of the Hungarian Academy of Sciences. 

\end{acknowledgement}



\bibliography{references.bib}



%
%
%


\end{document}